\documentclass[10pt,twocolumn,letterpaper]{article}

\usepackage[pagenumbers]{iccv} %

\definecolor{iccvblue}{rgb}{0.21,0.49,0.74}
\usepackage[pagebackref,breaklinks,colorlinks,allcolors=iccvblue]{hyperref}
\usepackage{multirow}

\usepackage{colortbl}
\usepackage{graphicx}
\usepackage{amsmath}
\usepackage{amssymb}
\usepackage{color}
\usepackage{booktabs}
\usepackage{enumitem}
\usepackage{tabularx} 
\usepackage{arydshln}
\usepackage{cuted}
\usepackage[table]{xcolor}
\definecolor{newlightblue}{RGB}{0,75,255}
\usepackage{array}
\usepackage[outline]{contour}
\usepackage{multirow}
\usepackage{xspace}
\usepackage[font=small]{caption}

\definecolor{baselinecolor}{gray}{.9}
\newcommand{\baseline}[1]{\cellcolor{baselinecolor}{#1}}
\newcommand{\modelname}{TrajViT\xspace}
\newcommand{\blname}{AutoMerge\xspace}

\def\paperID{1835} %
\def\confName{ICCV}
\def\confYear{2025}

\title{One Trajectory, One Token: \\ Grounded Video Tokenization via Panoptic Sub-object Trajectory}

\author{%
Chenhao Zheng$^{1,2}$, \
Jieyu Zhang$^{1,2}$, \
Mohammadreza Salehi$^{1,2}$, \
Ziqi Gao$^{1}$, \\
Vishnu Iyengar$^{1}$, \
Norimasa Kobori$^{3}$, \
Quan Kong$^{3}$, \
Ranjay Krishna$^{1,2}$ \
\\ 
\\
$^1$University of Washington, 
$^2$Allen Institute for Artificial Intelligence,
$^3$Woven by Toyota, Inc\\
}

\begin{document}
\maketitle

\begin{abstract}

Effective video tokenization is critical for scaling transformer models for long videos. Current approaches tokenize videos using space-time patches, leading to excessive tokens and computational inefficiencies. The best token reduction strategies degrade performance and barely reduce the number of tokens when the camera moves. 
We introduce \textit{grounded video tokenization}, a paradigm that organizes tokens based on panoptic sub-object trajectories rather than fixed patches. Our method aligns with fundamental perceptual principles, ensuring that tokenization reflects scene complexity rather than video duration. We propose \textit{\modelname}, a video encoder that extracts object trajectories and converts them into semantically meaningful tokens, significantly reducing redundancy while maintaining temporal coherence. 
Trained with contrastive learning, \modelname significantly outperforms space-time ViT (ViT3D) across multiple video understanding benchmarks, e.g.,  \modelname outperforms ViT3D by a large margin of 6\% top-5 recall in average at video-text retrieval task with 10x token deduction. We also show \modelname as a stronger model than ViT3D for being the video encoder for modern VideoLLM, obtaining  an average of 5.2\% performance improvement across 6 VideoQA benchmarks while having 4x faster training time and 18x less inference FLOPs. 
\modelname is the first efficient encoder to consistently outperform ViT3D across diverse video analysis tasks, making it a robust and scalable solution. 

\end{abstract}    
\section{Introduction}

\begin{figure}[t]
\centering
    \centering
    \raggedright
    \includegraphics[width=\columnwidth]{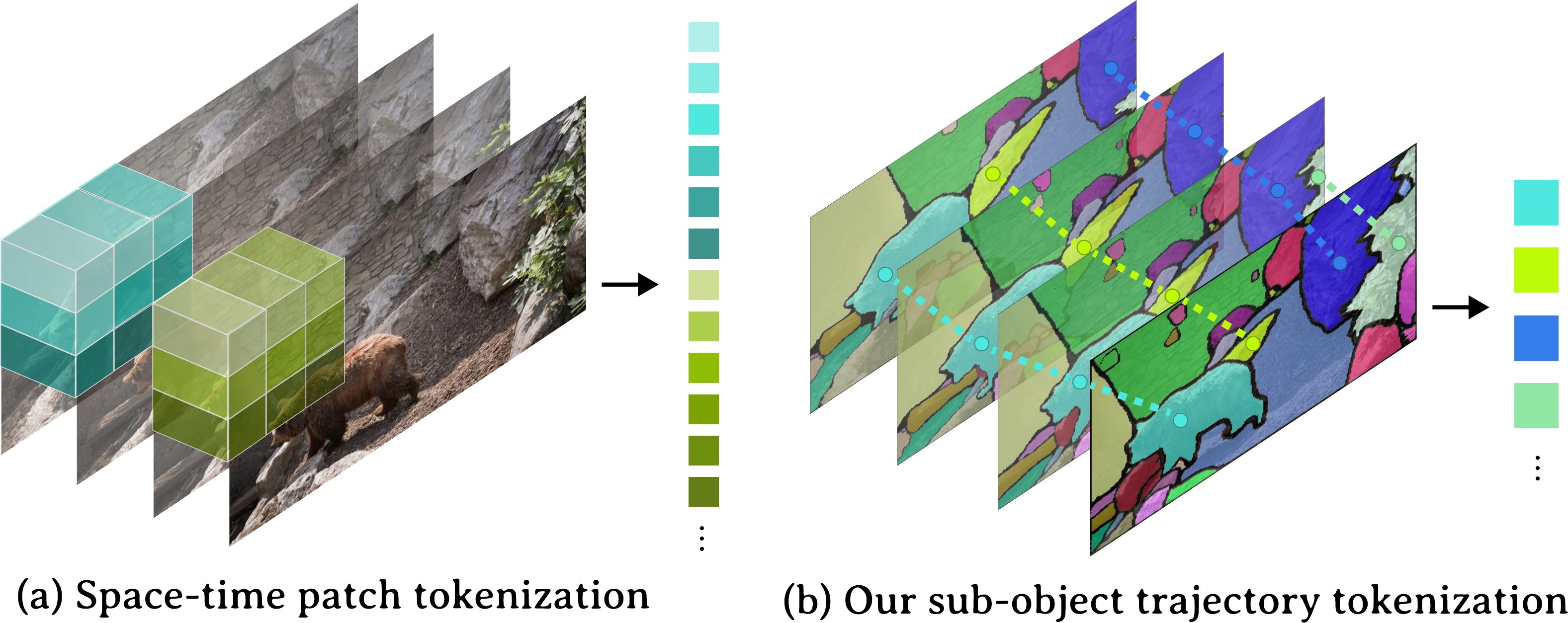}
    \caption{(a) Traditional video tokenization divides a video into space-time patches, leading to large number of redundant tokens. (b) We propose to represent a video via panoptic sub-object trajectory, where each token represents a trajectory.}
    \label{fig:teaser}
\end{figure}

With video comprising over 80\% of global internet traffic~\cite{cisco_vni_forecast}, designing efficient video encoders is essential. 
Since the transformer architecture~\cite{vaswani2017attention} is now the standard for encoding video data~\cite{neimark2021video,lin2022swinbert}, efficient design hinges on the tokenization algorithm.
Today's de-facto tokenization algorithm splits the video tensor into space-time patches~\cite{neimark2021video} as shown in Fig.\ref{fig:teaser}(a). In other words, the longer the video, the more tokens are extracted.

This tokenization paradigm introduces significant challenges: First, it leads to memory bottlenecks, as a one-minute video sampled  at just one frame per second could produce over 6000 tokens~\cite{arnab2021vivit}. Second, it does not respect the scene complexity, treating all videos equally regardless of their visual content. For instance, a one-minute video of a static office desk without any motion produces the same number of tokens as a one-minute video of a football match with high-speed athleticism. 
Third, it places a heavy burden on the transformer encoder to connect disjoint tokens across frames as a singular concept (\eg, the same office desk).
Researchers have proposed various token merging strategies to mitigate these challenges, but they are either not content-aware~\cite{ryoo2021tokenlearner,bolya2022token, arnab2021vivit} or perform poorly when the camera taking the video is moving~\cite{choudhury2025don}.

We depart from patch-based tokenization to propose a new paradigm of \textit{grounded video tokenization}. 
Instead of naively splitting videos into patches to be tokenized, we align tokenization with the fundamental perceptual principles governing object perception and motion~\cite{wagemans2012century,spelke1990principles, pylyshyn1988tracking} by organizing tokens to correspond to panoptic sub-object trajectories.  
Rooted in Spelke’s core cognitive principles~\cite{spelke1990principles} and the Gestalt Principle of \textit{common fate}, we treat object parts as coherent entities that persist over time (Fig.~\ref{fig:teaser}b). 
By correlating token count with the number of distinct panoptic sub-object trajectories—rather than frame count—our method directly reflects the complexity of a scene in terms of its constituent objects and their interactions, allowing efficient processing of long sequences. 
Moreover,
because our tokenization divides a video into objects rather than raw pixel divisions, it remains robust to variations in lighting conditions, occlusions, and camera movements—a challenge for patch-based tokenizers~\cite{buch2022revisiting, lei2022revealing, lei2021less}.

\begin{figure}[t]
    \centering
    \includegraphics[width=\linewidth]{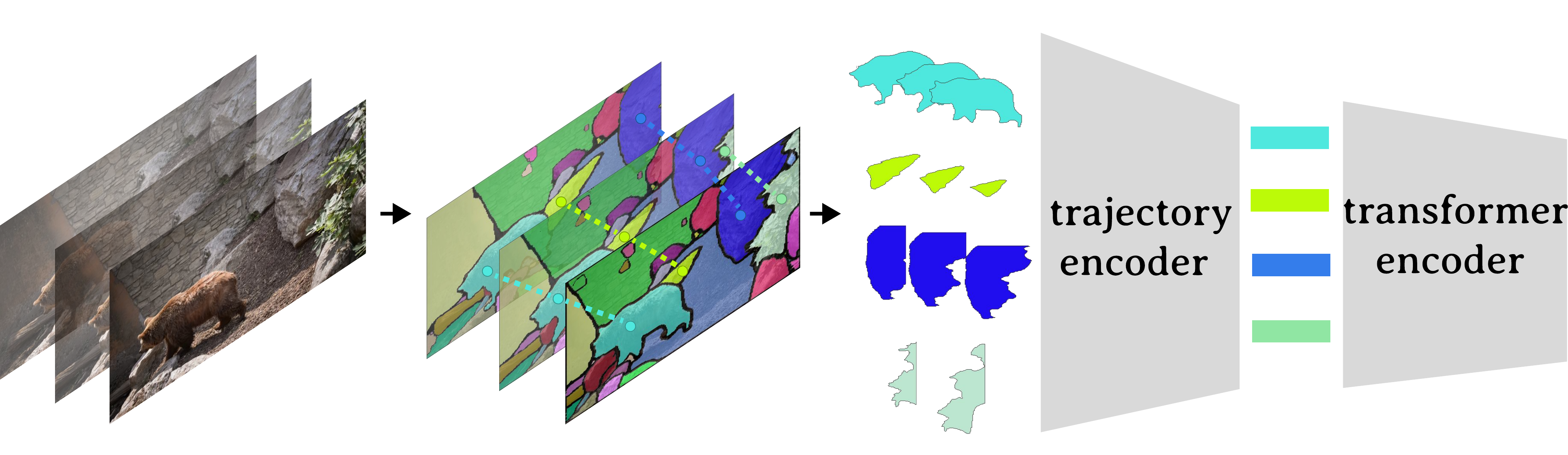}
    \caption{\textbf{Overview of \modelname.} Given a video, we first panoptically extract the trajectories for all objects. Our trajectory encoder converts these dynamic object trajectories into fixed sized embeddings, which serve as the input to the transformer encoder.}
    \vspace{-4mm}
    \label{fig:overview}
\end{figure}

We operationalize this paradigm by designing \textit{\modelname}, a video encoder with grounded video tokenization and a standard transformer encoder (Fig.~\ref{fig:overview}). 
Given a video, \modelname extracts panoptic trajectories for all object parts and converts each into a fixed-size token. All tokens are then processed by a transformer encoder. 
In addition to its content-aware nature, \modelname\ offers several advantages: (1) it significantly reduces the number of tokens—by an average of $10\times$ in the Panda dataset~\cite{chen2024panda}—compared to space-time patch tokenization.
(2) the tokens are organized into semantically meaningful units with no temporal redundancy, both of which facilitate high-level reasoning about object interactions for the encoder;
(3) it makes the tokenization invariant to low-level visual variations in lighting and camera motion, leading to improved performance in video understanding tasks despite using fewer tokens.

In designing \modelname, our core technical challenge lies in developing an efficient trajectory generation pipeline that can extract trajectories panoptically with sub-object granularity,  and a module that can encode distinct trajectories into fixed-size tokens.
We first develop a trajectory generation pipeline that integrates off-the-shelf segmenter~\cite{chen2024subobject} and tracker~\cite{ravi2024sam} through a parallelizable algorithm, which generates panoptic sub-object trajectories while minimizing latency caused by the tracking model.
We also design a lightweight \textit{trajectory encoder} that maps each distinct trajectories with variable number of pixels and appearing frames into fixed-size tokens, while maintaining the information about their appearance and temporal position variations. \modelname can also process image data by treating each image segment as a trajectory of length one, allowing seamingless joint training with both videos and images.
Finally, we train \modelname with a contrastive learning objective on a large-scale video/image-text dataset.

 We compare \modelname with standard ViT with space-time patch tokens (ViT3D) and state-of-the-art token merging methods on a wide range of video understanding tasks, including video-text retrieval, spatial-temporal detection, action classification, and temporal clip retrieval.
It outperforms VIT3D in all tasks, while all token merging baselines underperform ViT3D on average. At the largest training scale, \modelname surpasses ViT3D by a significant margin of 6\% in top-5 recall with 10x fewer tokens.
\modelname is particularly effective because most video tasks are object-centric or require fine-grained recognition.
For efficiency, despite the additional overhead of trajectory generation, \modelname trains faster, consumes less GPU memory, and performs faster inference for videos with more than 64 frames. 

Lastly, we train two VideoLLMs by connecting Llama3 with \modelname and ViT3D as video encoders. On six VideoQA benchmarks, the average accuracy of the \modelname-LLM surpasses ViT3D-LLM by 5.24\%, while being trained 4x faster and running at 18x fewer inference FLOPs. On the MovieChat benchmark~\cite{song2024moviechat} for long video evaluation, \modelname-LLM displays better accuracy and efficiency than ViT3D-LLM when we scale up the number of input frames.

\section{Related Work}

\noindent\textbf{Video tokenization.} In the field of video understanding, ViViT \cite{arnab2021vivit} introduced spatio-temporal patches to tokenize and process videos with transformers more efficiently than tokenizing video frames independently \cite{radford2021learning,luo2021clip4clipempiricalstudyclip}. Subsequent research has exploited the temporal redundancy of videos and improved tokenization efficiency through  patch dropping \cite{feichtenhofer2022maskedautoencodersspatiotemporallearners,choudhury2025don,zhang2025videollama3frontiermultimodal} or alternative input forms such as motion vector tokenization \cite{jin2024videolavitunifiedvideolanguagepretraining}. 
Recent researches like~\cite{choudhury2025don, fang2025sam2act} has made tokenization more content-aware with token counts adapting to video complexity, but performs poorly when the camera taking the video is moving.

\noindent\textbf{Efficient video encoders.} Prior work has improved video encoder efficiency through approaches like token merging or pruning in intermediate transformer layers \cite{bolya2022token,liang2022not,cao-etal-2023-pumer,choi2024vidtldrtrainingfreetoken,pan2021iared2interpretabilityawareredundancyreduction}, compressing content into fewer learnable tokens \cite{ryoo2021tokenlearner,li2023llamavidimageworth2,zhang2025llavaminiefficientimagevideo,attention_bottleneck} or via sampling tokens based on importance scores \cite{fayyaz2022adaptivetokensamplingefficient,wang2021efficient}. 
Unfortunately, the reduction in tokens is often accompanied by a reduction in model performance~\cite{arnab2021vivit, bolya2022token, choudhury2025don}.
Other works have introduced architectural changes, such as modifying the attention mechanism \cite{li2023svitt,shu2024videoxlextralongvisionlanguage} or implementing state-space models to improve video modeling efficiency \cite{li2024videomambastatespacemodel}. Our work focuses on token reduction without modifying transformer architecture.

\noindent\textbf{Efficient video large language models.}
In the context of large video language models (VideoLLMs), to reduce the number of video tokens in the language model's context length, previous research has proposed merging or dropping vision encoder output tokens \cite{chen2024imageworth12tokens,yang2024visionziplongerbetternecessary, rao2024unisoccer}. This is commonly accomplished via learnable components such as perceiver resampler \cite{jaegle2021perceiver,alayrac2022flamingo} or through instruction conditioned token pruning \cite{li2023blip2bootstrappinglanguageimagepretraining,shen2024longvu,li2024videochatchatcentricvideounderstanding,li2023llamavidimageworth2}. Notably, our approach is orthogonal to all these methods as we modify the tokenization before the transformer models.

\section{Grounded Video Tokenization}

\begin{figure}[t]
    \centering
    \includegraphics[width=\linewidth]{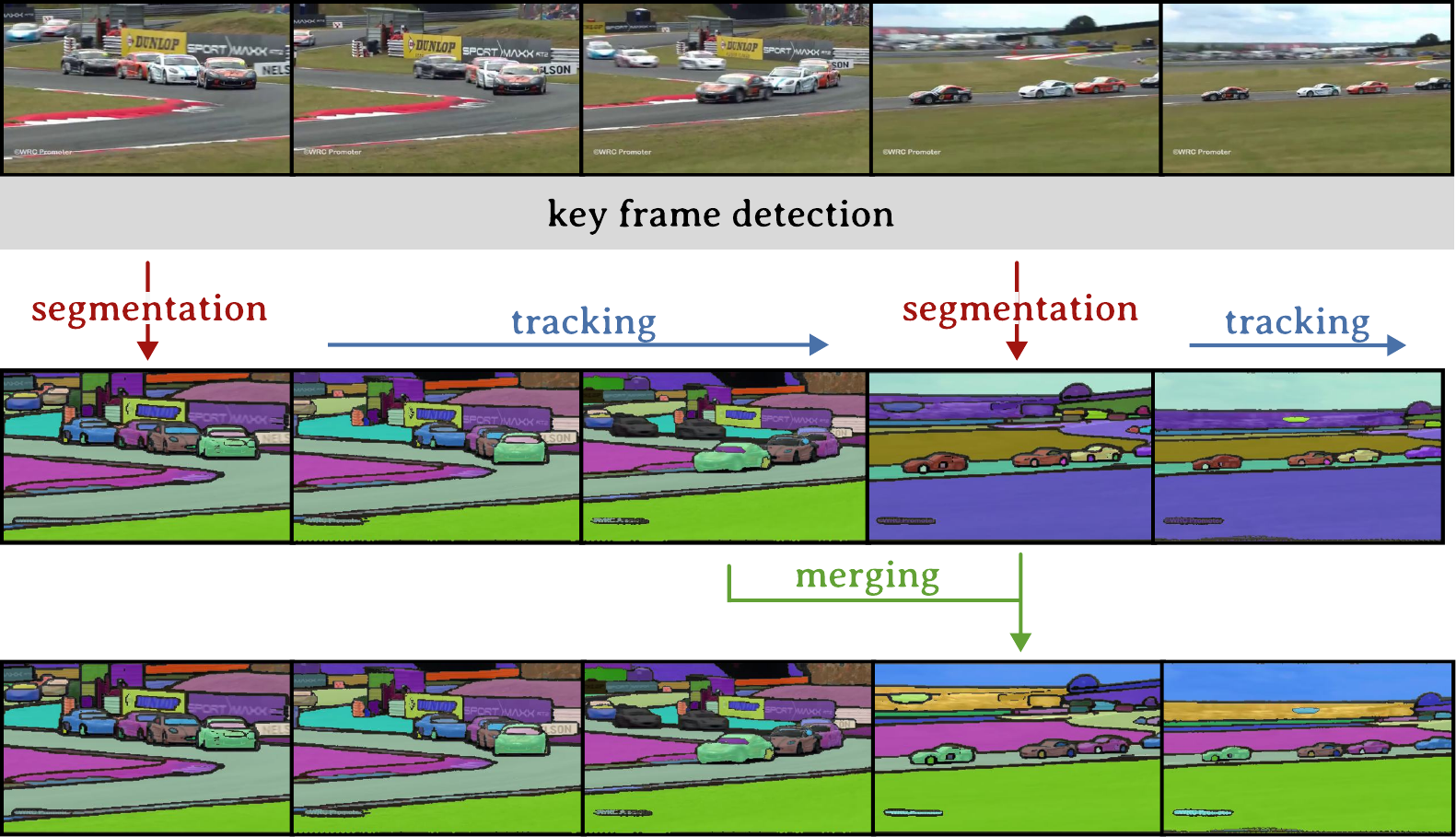}
    \caption{\textbf{Our parallel trajectory generation pipeline.} We use key frame detection to break a video into subclips. We segment and track objects in each clip in parallel and finally merge objects between clips. This paradigm captures objects that emerge over time while reducing overall tracking latency.}
    \label{fig:traj-gen}
\end{figure}
\begin{figure*}[htbp]
    \centering
    \includegraphics[width=\linewidth]{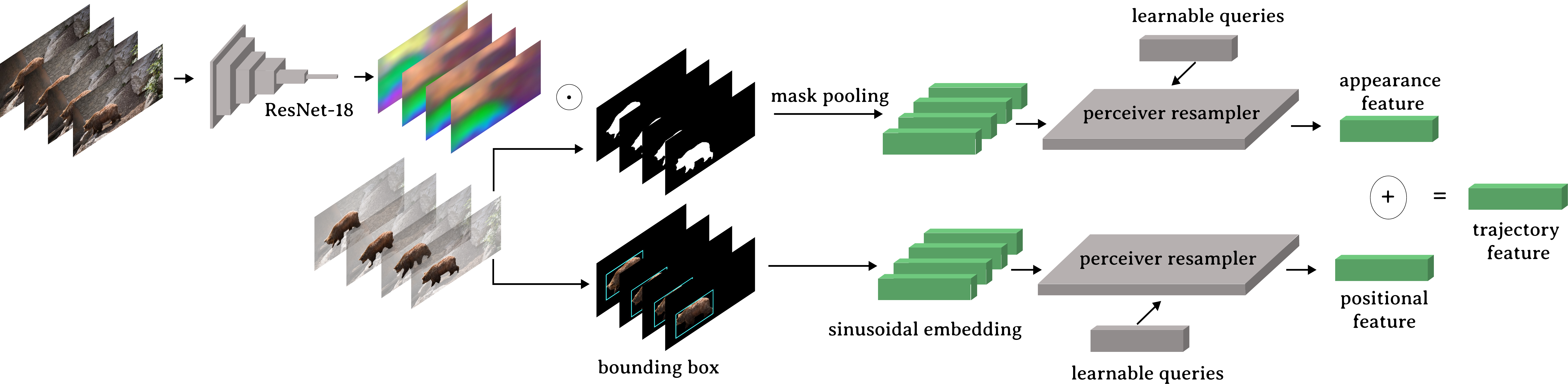}
    \caption{\textbf{Architecture of trajectory encoder.} we employ a two-branch design that enocodes a trajectory's appearance and temporal position separately. At each frame, we represent the appearance of a segment by mask pooling its feature, and represent its position by bounding box coordinates. Both features are then aggregated across frames via perceiver resampler and added together to form the  trajectory feature.}
    \vspace{-4mm}
    \label{fig:architecture}
\end{figure*}

\begin{figure}
    \centering
    \includegraphics[width=0.99\linewidth]{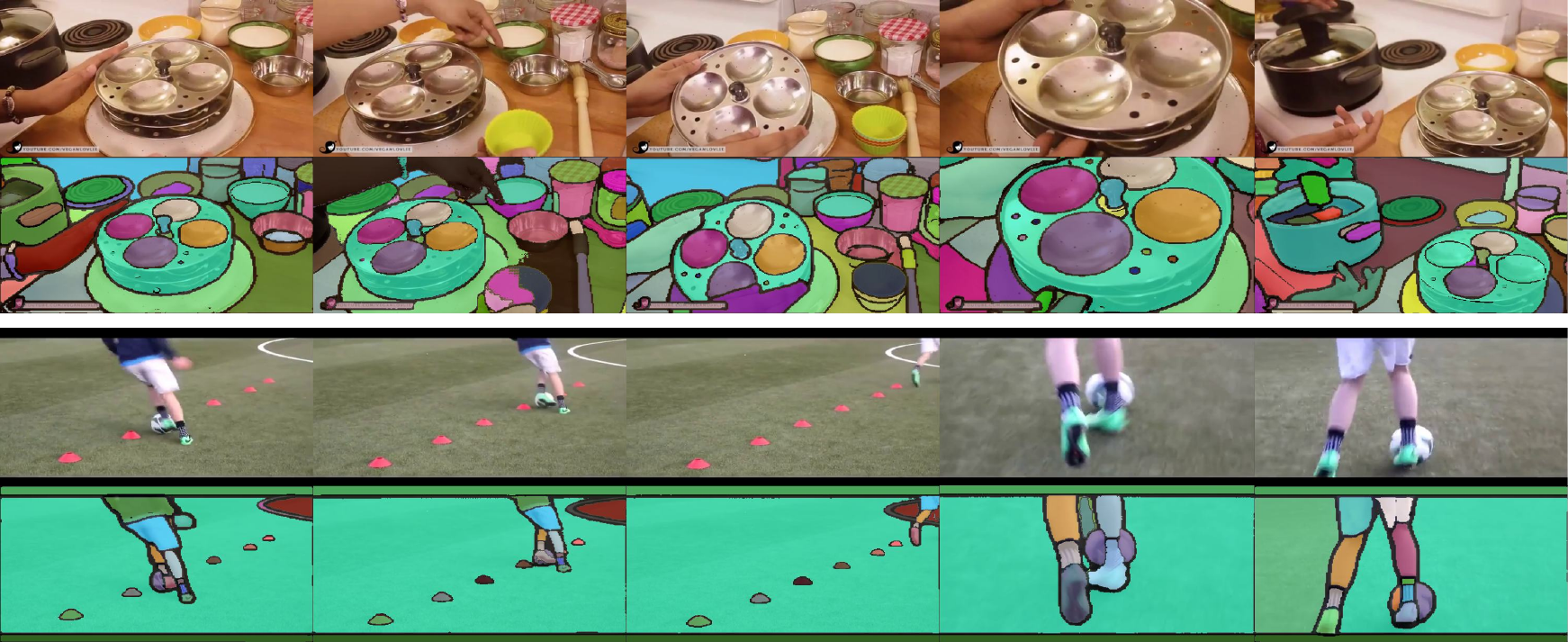}
    \caption{\textbf{Visualizations of generated trajectories.}}
    \label{fig:last_visual}
    \vspace{-4mm}
\end{figure}

Our approach aims to represent videos using semantically meaningful and compact tokens derived from panoptic sub-object trajectories.
As illustrated in Fig.~\ref{fig:overview}, we panoptically extract trajectories for all objects in a video and convert them into fixed-size embeddings to be used as the transformer encoder input. We find that trajectory extraction and encoding each pose distinct challenges. First, existing tracking solutions, which propagate segmentation masks from frame to frame, cannot handle newly appearing objects mid‐video and are prone to long propagation time~\cite{ravi2024sam, yang2023track, cheng2024putting}, necessitating a new trajectory generation pipeline. Second, the trajectories we generated can contain a varying number of pixels and appear in a varying number of frames, but the final representations need to be fixed-size embeddings to be compatible with the transformer architecture, requiring a new trajectory encoding module. We detail how we tackle these challenges in this section.

\subsection{Panoptic Sub-object Trajectory Extraction}
\label{subsec:traj}
Our objective is to accurately segment and track all objects in a video—including backgrounds—while keeping the process efficient. We design a pipeline that couples an off‐the‐shelf segmentation and tracking model, can capture objects that emerge over time, and can be parallelized to reduce latency. 

\noindent\textbf{Discovering objects by segmenting key frames.} Given a video of length $T$ with time index $t \in \{1, \dots, T\}$, we begin by selecting key frames $\{\hat{t_i}\}_{i=1}^K$ where new objects are likely to appear. Specifically, we compare the color histogram and luminance intensity between consecutive frames and flag those exceeding a manual threshold (details in supplementary). Then, we panoptically segment each key frame to capture all objects. We employ the DirectSAM model~\cite{chen2024subobject} as it is lightweight with 84M parameters and can produce fine-grained segments in one forward pass with consistent granularity across images, reducing the risk of merging failure in the later merging stage. 
We denote the segmentation mask derived in key frames as $\{M_{\hat{t_i}}\}_{i=1}^K$.

\noindent\textbf{Tracking objects within each clip.}
The detected key frames naturally partition the video into multiple clips. Within each clip 
$\hat{t_i} \leq t < \hat{t_{i+1}}$,
we employ an off‐the‐shelf tracker (SAM2~\cite{ravi2024sam}) to propagate the segmentation mask  $M_{\hat{t_i}}$ from the key frame $\hat{t_i}$ to subsequent ones. This per‐clip process can be executed in parallel for a batch of clips for efficiency (Fig.~\ref{fig:traj-gen}). In practice, we additionally split a clip at the midpoint if its length exceeds a certain threshold to shorten the propagation time. We use hiera-small variants of SAM2 with 46M parameters. 

\noindent\textbf{Merging objects between consecutive clips.} Finally, we link the identities of objects between consecutive clips. 
Specifically, we perform an additional tracking step from the final frame of a clip ($\hat{t_{i+1}} - 1$) to the first frame of the subsequent clip ($\hat{t_{i+1}}$), generating a propagated mask $M_{\hat{t_{i+1}}}'$. We 
then compare each segment in $M_{\hat{t_{i+1}}}'$ with those in the original key frame mask $M_{\hat{t_{i+1}}}$. 
If the Intersection over Union (IoU) for any pair of segments exceeds a threshold of 0.8, we 
consider them to represent the same object and merge their trajectories.

This split-track-merge pipeline efficiently discovers and tracks objects across entire videos. The visualizations of generated trajectories are shown in supplementary. While we observe occasional inconsistent tracking, i.e., segments of the same object in consecutive clips being tagged as different trajectories due to imperfect merging, it seldomly hurts model performance because we find that these inconsistencies often coincide with situations where the object undergoes a large appearance change. In such cases, creating a new trajectory might benefit downstream tasks, as it accounts for significant changes in objects.

\subsection{Video Encoding with Trajectory Tokens}
\label{subsec:tokenizer}
We aim to obtain fixed-size representations of trajectories and utilize them as input for a transformer model.
To achieve this, we develop a \emph{trajectory encoder} that produces an appearance embedding and a temporal position embedding for each trajectory (Fig.~\ref{fig:architecture}).

\noindent\textbf{Trajectory appearance encoding.} 
For trajectory appearance encoding, we first extract hierarchical visual features from each frame. Given a video consisting of frames $\{I_t\}_{t=1}^T$, we apply a lightweight convolutional network~\cite{he2016deep} to obtain corresponding hierarchical feature maps $\{F_t\}_{t=1}^T$ for each frame. Then, for each trajectory, we extract its appearance feature via masked feature pooling. Specifically, for trajectory $i$ at frame $t$, we compute:
\vspace{-4pt}
\begin{equation}
    f_t^i = \frac{\sum_{x,y} M_t^i(x,y) \cdot F_t(x,y)}{\sum_{x,y} M_t^i(x,y) + \epsilon}
\end{equation}
where $x,y$ denotes the spatial position, $M_t^i$ is the trajectory's segmentation mask, $F_t$ is the feature map, and $\epsilon$ is a small constant. This operation pools the feature inside the region-of-interest for each trajectory. The resulting $f_t^i$ represents the trajectory $i$’s feature at frame $t$.
Since a trajectory can span a varying number of frames, we need a mechanism to summarize a dynamic sequence of the per-frame features into a fixed-size embedding. We choose to use one layer of perceiver resampler module~\cite{alayrac2022flamingo}, which is a lightweight cross-attention layer that learns a predefined number of latent input queries to attend to the sequence of per-frame features. As we set the number of latent queries to be one, the perceiver resampler outputs a single embedding that serves as the appearance representation of the trajectory. We use rotary embedding~\cite{su2024roformer} as the positional embedding of perceiver resampler module for better test-time frame number generalization.

\noindent\textbf{Trajectory temporal position encoding.} 
To represent the temporal position information of each trajectory, we extract bounding box coordinates $c_t^i = (x_1^i, y_1^i, x_2^i, y_2^i)$ at each frame of the trajectory $i$.
These coordinates are then mapped to high-dimensional vectors using sinusoidal embedding~\cite{mildenhall2021nerf}. Since the bounding box sequence for a trajectory is also of variable length, we again employ the perceiver resampler module to summarize the sequence into a single temporal position embedding. 

Finally, we construct the trajectory embedding by summing the appearance embedding and the temporal position embedding, which serves as a single token of the input to the transformer.
The trajectory encoder has 20M parameters, much smaller than the transformer encoder (e.g., 304M for ViT-Large), yet can significantly reduce the number of tokens for efficient video processing, without dropping information in raw pixels.

\noindent\textbf{Model Training.}
\label{subsec:pretraining}
Our final video encoder \textit{\modelname} consists of this trajectory encoder followed by a standard transformer module. 
We train it with the CLIP objective~\cite{radford2021learning}.
Given a dataset of video-caption pairs, CLIP objective learn video and text representations by jointly training a video encoder and a text encoder with contrastive loss.

\begin{table*}[h]
\centering
\setlength{\tabcolsep}{8pt} %

\begin{minipage}{0.63\textwidth} %
    \centering
    \small %
    \centering
\setlength{\tabcolsep}{3pt}
\resizebox{.8\linewidth}{!}{\begin{tabular}{lcccccccc}
    \toprule
    \multirow{2}{*}{\textbf{Model}} & 
    \multicolumn{2}{c}{\textbf{ActivityNet}} & 
    \multicolumn{2}{c}{\textbf{VATEX}} &
    \multicolumn{2}{c}{\textbf{MSR-VTT}} &
    \multicolumn{2}{c}{\textbf{Charades}} \\
    
    \cmidrule(lr){2-3} \cmidrule(lr){4-5} \cmidrule(lr){6-7} \cmidrule(lr){8-9}
    & {\small txt2vid} & {\small vid2txt} & {\small txt2vid} & {\small vid2txt} & {\small txt2vid} & {vid2txt} & {\small txt2vid} & {\small vid2txt} \\ 
    \midrule
    ViT3D & 35.3 & 34.5 & \underline{35.0} & \underline{59.1} & \underline{30.9} & 56.5 & 12.9 & 12.6 \\
    TokenLearner & \underline{36.9} & \underline{36.8} & 34.1 & 58.5 & 30.3 & \underline{58.7} & 11.7 & \underline{12.8} \\
    ViViT & 33.4 & 32.9 & 34.0 & 57.9 & 29.5 & 54.6 & 12.0 & 12.3 \\
    \blname & 34.0 & 34.2 & 34.2 & 58.1 & 29.1 & 55.9 & 10.3 & 11.4 \\
    RLT & 33.6 & 33.3 & 34.1 & 58.4 & 30.1 & 56.1 & \underline{13.4} & \underline{12.8} \\
    ToMe & 31.4 & 31.4 & 34.0 & 56.6 & 28.6 & 55.31 & 10.2 & 10.5\\
    \modelname (ours) & \textbf{38.6} & \textbf{38.4} & \textbf{36.2} & \textbf{61.0} & \textbf{31.7} & \textbf{61.0} & \textbf{14.9} & \textbf{14.8} \\
    \bottomrule
\end{tabular}}
\vspace{-2mm}
\caption{\textbf{Zero-shot video-text retrieval performance.} We report R@5 scores in four commonly used video retrieval datasets. Txt2vid stands for text-to-video while vid2txt stands for video-to-text. Our model consistently surpass all baselines by a large margin.}
\label{tab:retrieval}

\end{minipage}
\hfill
\begin{minipage}{0.33\textwidth} %
    \centering
    \small %
    \centering
\setlength{\tabcolsep}{4pt}
\resizebox{.8\linewidth}{!}{\begin{tabular}{lccc} 
    \toprule
    \textbf{Model} & \textbf{K400} & \textbf{SSV2} & \textbf{UFC-101} \\
    \midrule
    ViT3D & \underline{53.4} & \textbf{46.1} & \underline{83.8} \\
    TokenLearner & 53.1 & 42.7 & 83.5 \\
    ViViT & 51.0 & 43.0 & \underline{83.8} \\
    \blname & 49.3 & 41.2 & 83.3 \\
        RLT & 53.1 & 44.5 & 83.5\\
    ToMe & 50.3 & 41.7 & 82.9 \\ 
    \modelname (ours) & \textbf{55.6} & \underline{45.8} & \textbf{84.5} \\
    \bottomrule
\end{tabular}}
\vspace{-2mm}
\caption{\textbf{Action classification performance.} Scores are reported in terms of top-1 accuracy. Our model is the only token merging method that is comparable with ViT3D's performance.}
\label{tab:classification}

\end{minipage}
\vspace{-4mm}
\label{tab:combined}
\end{table*}

\begin{table}[h]
\centering
\small
\setlength{\tabcolsep}{3pt}
\resizebox{.8\linewidth}{!}{\begin{tabular}{lccccc} 
    \toprule
    \multirow{2}{*}{\textbf{Model}} & 
    \textbf{AVAv2} & 
    \multicolumn{2}{c}{\textbf{ActivityNet}} & 
    \multicolumn{2}{c}{\textbf{YouCook}}  \\
    
    \cmidrule(lr){2-2} \cmidrule(lr){3-4} \cmidrule(lr){5-6} 
     & {\small mAP} & {\small txt2vid} & {\small vid2txt} & {\small txt2vid} & {\small vid2txt} \\ 
    \midrule
    ViT3D & \underline{10.6} & \underline{39.3} & 42.3 & \underline{46.8} & \textbf{50.0} \\
    TokenLearner & 9.1 & 39.1 & \textbf{43.0} & 46.0 & 49.8  \\
    ViViT & 9.9 & 38.9 & 42.3 & 45.7 & 48.8\\
    \blname & - & 38.9 & 41.8 & 45.7 & 47.8\\
    RLT & 8.5 & \underline{39.3} & 42.1 & 46.2 & 49.4\\
    ToMe & 8.3 & 38.4 & 42.0 & 45.4 & 48.2\\
    \modelname (ours) & \textbf{13.7}  & \textbf{39.6} & \underline{42.5} & \textbf{47.2} & \textbf{50.0} \\
    \bottomrule
\end{tabular}}
\vspace{-2mm}
\caption{\textbf{Fine-grained video tasks.}  AVAv2 is for spatial-temporal action detection task. ActivityNet and Youcook are for temporal clip retrieval task. }
\label{tab:fine_grained}
\end{table}

\subsection{Seamless Integration with Image Data}
\label{subsec:image}
Incorparating image data is a common practice when training video encoders, as the size of publicly available image data is several orders of magnitude larger than that of video data.
\modelname can integrate image data naturally without architectural modifications. Given an image, we apply the same segmentation process as in videos, and then treat each segmented region as a trajectory of length one. This enables seamless integration of image data into video pretraining.
In contrast, models that use space-time patch tokenization cannot directly process image data because of their reliance on 3D convolutions or MLP layers that map space-time patches to fixed-size tokens. These modules require channel dimensions that match the spatial-temporal patch size, preventing direct image processing. As a result, space-time patch based approaches typically pretrain ViT2D on images, transfer the weights to ViT3D, and then fine-tune on video data, while ours allows joint training on images and videos in one stage.

\section{Experiments}

We evaluate \modelname on a diverse set of video understanding tasks. Our experiments cover three key areas:
(1) general video understanding tasks that can be assessed directly using our trained CLIP model (Sec.~\ref{sec:gen_video_understand});
(2) model scaling behavior when the number of frames during inference or pretraining dataset size increases, as well as when image data is added to the pretraining corpus (Sec.~\ref{sec:scaling}); and
(3) video-language QA tasks that can be evaluated by connecting a video encoder to an LLM (Sec.~\ref{sec:qa}). 
We also include an ablation study in Sec.~\ref{sec:ablation}. 

\noindent\textbf{Training recipe.}  
Models in the main experiments use the transformer architecture of ViT-Large~\cite{dosovitskiy2020image} without pre-trained weights. During training, we uniformly sample a 16-frame clip from each video. 
All models are trained on a randomly sampled subset of the Panda-70M video captioning dataset~\cite{chen2024panda}. We partition the dataset into three scales: 2M, 4M, and 8M video samples. \modelname and ViT3D are trained on all three scales to analyze scaling performance, while all other baselines are trained on the 4M subset due to computational constraints. In one of our experiments that adds image data into pretraining, we use 50M randomly sampled image-text pairs from DataComp-1B~\cite{gadre2023datacomp} as the image dataset.

\noindent\textbf{Baselines.}  We compare \modelname with following baselines in a strict apple-to-apple manner, ensuring they share the same training dataset and recipe.
\begin{itemize}
    \item \textbf{Space-time ViT (ViT3D):} A widely used baseline that tokenizes videos using fixed space-time patches and processes them with a standard transformer encoder~\cite{arnab2021vivit}.
    \item \textbf{\blname:} 
    A variant of our \modelname which uses a learnable token merging mechanism on top of space-time patches to replace trajectory tokens. This serves as a control to study the impact of trajectory tokens (architecture detail in supplementary). 
    We adjust its produced token count to match its FLOPS with that of \modelname at 16-frame videos in the training set for a fair comparison.
    \item \textbf{Video token merging methods:} We compare with state-of-the-art token merging encoders such as ToMe~\cite{bolya2022token}, RLE~\cite{choudhury2025don}, and TokenLearner~\cite{ryoo2021tokenlearner},  which reduce video token counts by dropping similar patches or intermediate patch features. We follow the default token reduction hyperparameters from the original papers for all methods.
    \item \textbf{Methods that modify transformer architectures:} We additionally include a widely-used video architecture ViViT~\cite{arnab2021vivit}, which improves video processing efficiency by modifying transformer design. 
\end{itemize}

\subsection{General Video Understanding Tasks}
\label{sec:gen_video_understand}

\begin{figure*}[t]
    \centering
    \includegraphics[width=\linewidth]{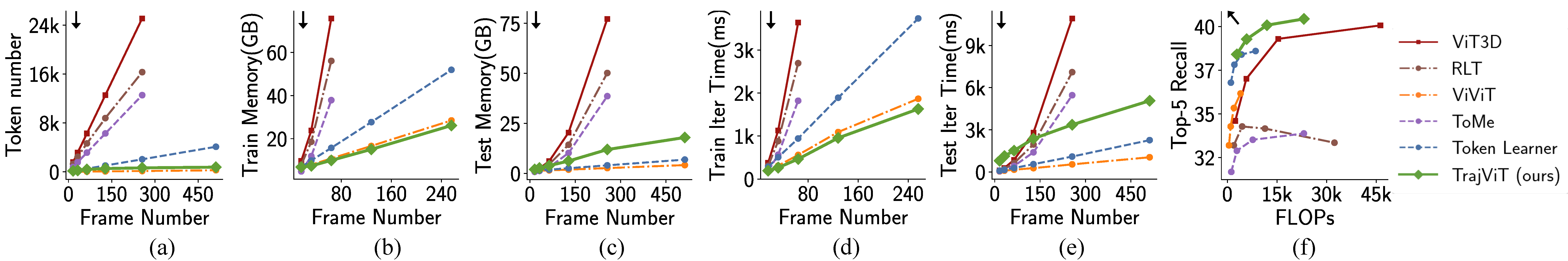}
    \caption{\textbf{Comparison of inference frame number scaling in ActivityNet video-to-text retrieval task.} Scaling with our tokenization paradigm obtains a better trade-off than baselines in terms of efficiency and accuracy.}
    \label{fig:frame_scaling}
\end{figure*}
\begin{figure}[tbp]
    \centering
    \begin{minipage}{0.48\linewidth} %
        \centering
        \small %
        \includegraphics[width=\linewidth]{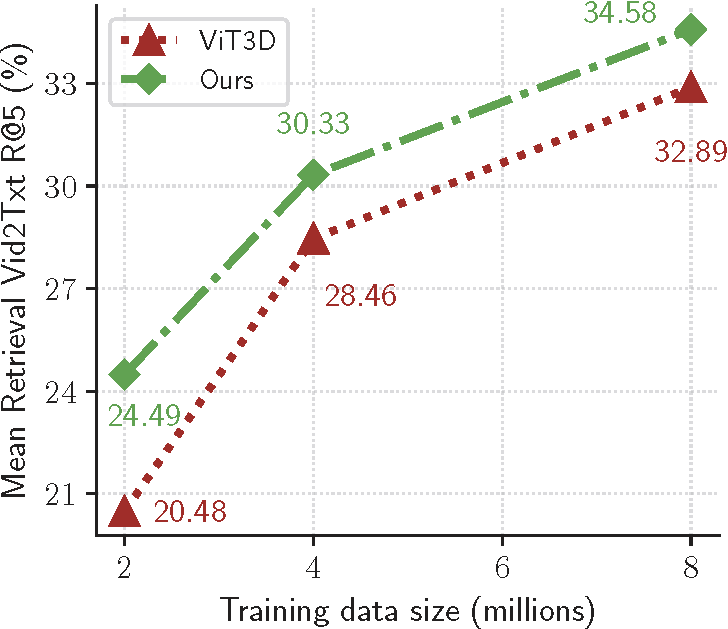}
        \caption{\textbf{Scaling video training data.} \modelname maintains the lead over ViT3D across different scales of training data.}
        \label{fig:scale_data}
    \end{minipage}
    \hfill
    \begin{minipage}{0.48\linewidth} %
        \centering
        \small %
        \includegraphics[width=\linewidth]{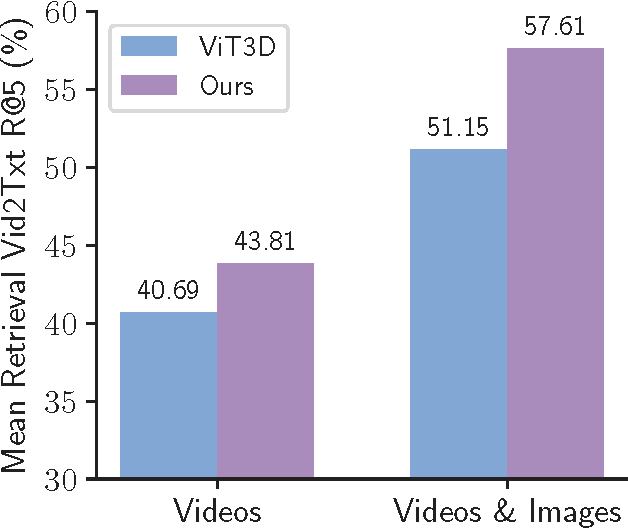}
        
        \caption{\textbf{Incorporating image data.}  \modelname benefits more from adding image data into pretraining as it requires no architectural modifications.}
        \label{fig:add_image}
    \end{minipage}
\end{figure}

We first compare \modelname against all baselines on a wide spectrum of general video understanding tasks that can be assessed by CLIP training, including action classification, video-text retrieval, and fine-grained spatial-temporal understanding. All models are trained on the 4M subset. We freeze the video encoders for all downstream evaluations. Unless otherwise specified, we uniformly sample 16 frames from videos as model input.

\noindent\textbf{Video-text retrieval.}  
We evaluate \modelname on multiple video-text retrieval benchmarks, including ActivityNet-caption~\cite{krishna2017dense}, Charades~\cite{sigurdsson2016hollywood}, MSR-VTT~\cite{xu2016msr}, and VATEX retrieval~\cite{wang2019vatex}. The evaluation is conducted in a zero-shot manner using our pretrained video and text encoder. As shown in Tab.~\ref{tab:retrieval}, \textbf{\modelname achieves a significant improvement over all baselines}. We attribute this to the nature of video-text retrieval tasks, where textual descriptions primarily focus on objects and their interactions. By leveraging semantic tokens derived from object trajectories, \modelname is better at identifying and matching objects, actions, and their relationships with the textual queries.

\noindent\textbf{Action classification.}  
We evaluate on three standard action recognition benchmarks: Something-Something V2 (SSV2)~\cite{goyal2017something}, Kinetics-400 (K400)~\cite{kay2017kinetics}, and UFC-101~\cite{soomro2012ucf101}.  Following prior works~\cite{wang2024internvideo2, zhao2024videoprism, yuan2023videoglue}, we employ a multi-head attention probing (MAP) head to probe the output representations from video encoders. We also follow standard practice to aggregate predictions from 4 temporal views per video~\cite{arnab2021vivit, yuan2023videoglue, wang2024internvideo2}. The results are presented in Tab.~\ref{tab:classification}. \textbf{\modelname outperforms all baselines on K400 and UCF-101}. On SSV2, our method significantly surpasses token merging approaches but slightly underperforms ViT3D. We observe that it might be because \modelname generates very few tokens in SSV2 due to the relatively simple egocentric scene, which forces the model to over-compress complex motion in a single token.

\noindent\textbf{Fine-grained spatial-temporal understanding.}  
To assess \modelname's capability in recognizing fine-grained details in a video, we evaluate on spatial-temporal detection task at AVAv2 benchmark~\cite{gu2018ava}, as well as temporal clip retrieval task of ActivityNet~\cite{krishna2017dense} and YouCook~\cite{das2013thousand} benchmarks. For spatial-temporal detection, we extract features from the region-of-interest and apply a MAP head to probe the action class (details in the supplementary) for each video encoder. For temporal clip retrieval, we follow ~\cite{zhao2024videoprism} to repurpose ActivityNet and YouCook for clip-to-text retrieval by segmenting each video into multiple clips based on the annotated timestamps. The retrieval task is then performed by selecting the correct description for a given video clip from the set of sequential descriptions within the same video. As shown in Tab.~\ref{tab:fine_grained}, \textbf{\modelname outperforms all baselines by a large margin in spatial-temporal detection tasks}. We attribute this to the fact that our trajectory tokens maintain object consistency over time, leading to more precise spatial and temporal grounding.
In temporal clip retrieval tasks, the performance gap between models is relatively small, possibly due to the presence of hard negatives that make the tasks challenging. However, \modelname still ranks highest in most metrics, demonstrating its ability to effectively differentiate video segments.

\begin{table*}[thbp]
\centering
\small
\setlength{\tabcolsep}{4pt}
\renewcommand{\arraystretch}{1.1}
\resizebox{.8\linewidth}{!}{\begin{tabular}{l cc cccc c cc cc c}
    \toprule
    \multirow{2}{*}{\textbf{Model}} & 
    \multicolumn{2}{c}{\textbf{NextQA}} & 
    \multicolumn{4}{c}{\textbf{TempCompass}} &
    \multicolumn{1}{c}{\textbf{VideoMME}} &
    \multicolumn{2}{c}{\textbf{ActivityNetQA}} &
    \multicolumn{2}{c}{\textbf{MovieChat}} &
    \multicolumn{1}{c}{\textbf{MLVU}} \\
    
    \cmidrule(lr){2-3} \cmidrule(lr){4-7} \cmidrule(lr){8-8} \cmidrule(lr){9-10} \cmidrule(lr){11-12} \cmidrule(lr){13-13}
    & OE & MC & MC & Yes/No & Cap.Match & Cap. & Acc. & Acc. & Score & Acc. & Score & Acc. \\
    \midrule
    ViT3D      & 49.2 & 29.5 & 39.1 & 50.1 & 58.4 & 30.4 & 28.9 & 30.7 & 2.6  & 34.7 & \textbf{3.1} & 19.2 \\
    \modelname & \textbf{65.1} & \textbf{37.0} & \textbf{40.0} & \textbf{50.6} & \textbf{59.9} & \textbf{31.1} & \textbf{32.0} & \textbf{38.0} & \textbf{2.7} & \textbf{36.7} & \textbf{3.1} & \textbf{32.2} \\
    \bottomrule
\end{tabular}}
\caption{\textbf{Comparison of ViT3D and \modelname on VideoLLM QA tasks.} We report results on NextQA (WUPS score in open-ended subsets, accuracy in multi-choice subsets), TempCompass (accuracy for four subsets of multi-choice, yes/no, caption matching, and captioning), VideoMME (accuracy), MovieChat (accuracy and GPT score), ActivityNetQA (accuracy and GPT score), and MLVU (accuracy). The VideoLLM that uses our video encoder consistently outperforms the one with ViT3D across all video QA benchmarks. }
\label{tab:videollm}
\vspace{-4mm}
\end{table*}

\subsection{Scaling Performance}
\label{sec:scaling}
We evaluate the scaling behavior of \modelname along three axes: number of input frames, pretraining data size, and joint training with images.

\noindent\textbf{Scaling the number of input frames.} 
We examine how different models perform as the number of input frames $T$ increases. We use the ActivityNet dense captioning retrieval task as our benchmark. This task requires retrieving long video clips ranging from 1 to 2 minutes, where incorporating more than 16 frames can help the model better align with the provided complex and densely annotated captions. We analyze both model performance and computational efficiency under varying temporal input lengths. 
For \modelname, we note that generating trajectories is a one-time cost for a video: we precompute and store trajectories used for every training epoch. At inference, we consider both the trajectory generation and the model forward pass as part of \modelname's resource consumption.

We show results in Fig.~\ref{fig:frame_scaling}. Compared to ViT3D, \modelname scales effectively to a significantly larger number of frames while maintaining low resource consumption. 
We observe that \modelname can generalize to more input frames with a smaller number of tokens (Fig.~\ref{fig:frame_scaling}a), lower training GPU memory (Fig.~\ref{fig:frame_scaling}b), and faster training iteration time (Fig.~\ref{fig:frame_scaling}d). When taking trajectory generation overhead into account at inference,  \modelname runs faster only for videos with more than 64 frames (Fig.~\ref{fig:frame_scaling}e), yet it consumes less GPU memory regardless of the number of frames (Fig.~\ref{fig:frame_scaling}c). Most importantly, Fig.~\ref{fig:frame_scaling}f demonstrates that \textbf{under the same computational cost, our approach consistently achieves better performance compared to all baselines}. These results highlight the potential of our design in building an efficient and strong long-term video understanding model. Note that the inference runtime and FLOPs of \modelname are bottlenecked by the tracking pipeline, which is a potential direction to improve in our future work.

\noindent\textbf{Scaling pretraining video data size.}  
To examine how \modelname scales with data, we train and compare \modelname and ViT3D on 2M, 4M, and 8M video subsets of Panda70M. In Table~\ref{fig:scale_data}, we report the average video-to-text retrieval performance across all four retrieval benchmarks from Sec. \ref{sec:gen_video_understand}. \textbf{\modelname consistently outperforms ViT3D across all three scales}, demonstrating equally strong scaling behavior. Interestingly, the performance gain at 2M is slightly larger than at 4M and 8M, possibly because the inductive biases introduced by our tokenization scheme provide greater benefits in lower-data regimes.

\noindent\textbf{Incorporating image data.}  
Most video encoders are initialized from image-pretrained models~\cite{arnab2021vivit, choudhury2025don, wang2024internvideo2, zhao2024videoprism, lei2022revealing}. We investigate whether \modelname can also benefit from image data. Unlike ViT3D, which requires separate image pretraining before video adaptation, \modelname can be jointly trained on image and video data as discussed in Sec.~\ref{subsec:image}. We conduct an experiment using the largest 8M video subset alongside the DataComp-50M image dataset~\cite{gadre2023datacomp}. We plot the average retrieval performance in Figure~\ref{fig:add_image}. Compared to ViT3D, \textbf{\modelname\ achieves a larger performance boost when incorporating image data}, highlighting its ability to leverage image and video data more efficiently.

\subsection{VideoLLM for QA tasks}
\label{sec:qa}
\begin{figure}[tbp]
    \centering
    \includegraphics[width=0.98\linewidth]{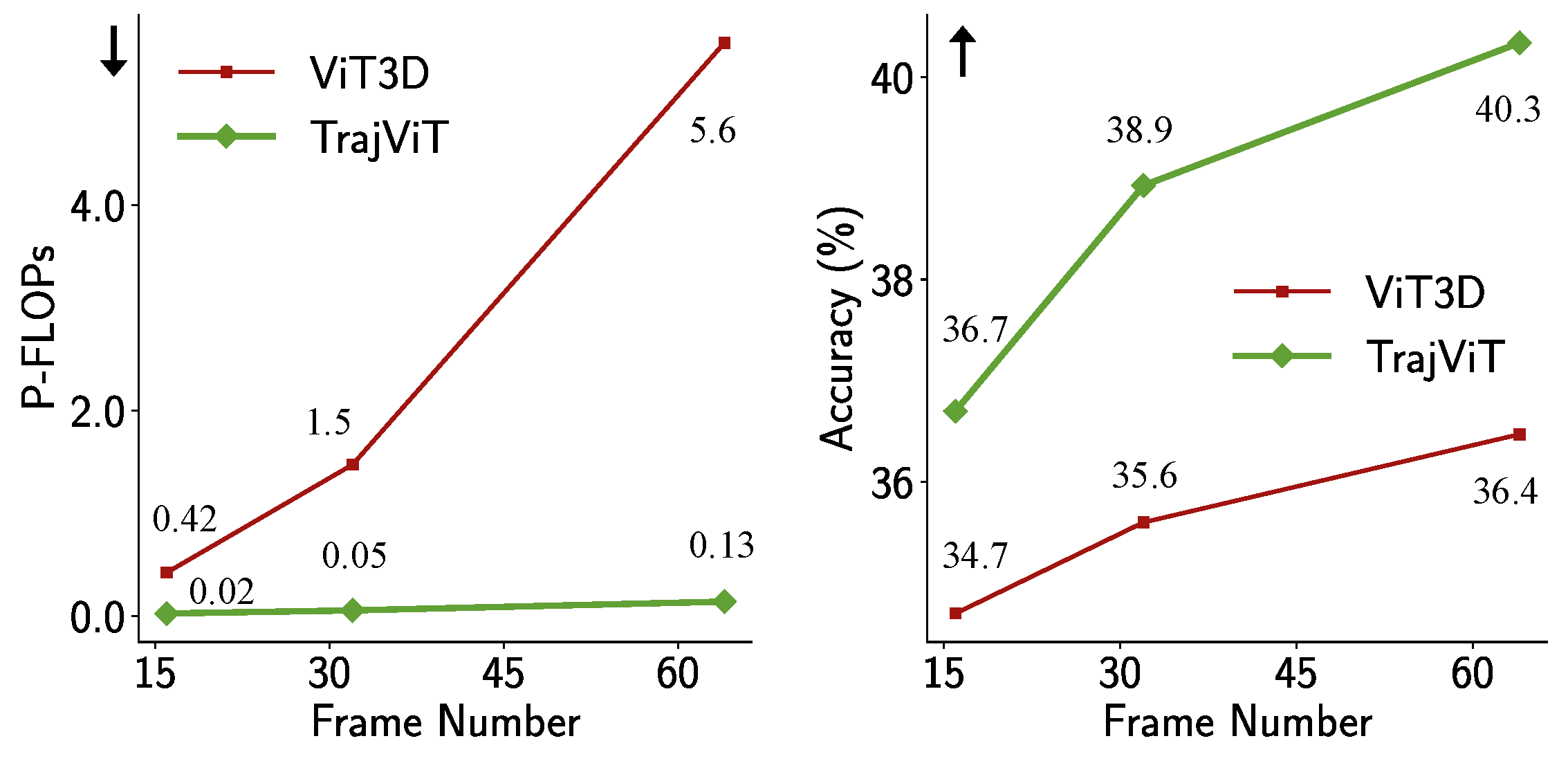}
    \caption{\textbf{Accuracy and inference FLOPs at MovieChat long video benchmarks with input frame scaling.} The VideoLLM with \modelname as video encoder scales significantly better than the one with ViT3D in both accuracy and efficiency.}
    \label{fig:vlm_scaling}
\end{figure}

We evaluate various VideoQA tasks by connecting \modelname to the Llama-3 language model~\cite{liu2024ppllava}. To focus on the impact of video encoders, we adopt the simple setup in~\cite{maaz2023video}: A linear layer is used to connect the trained video encoder and LLM. We train two VideoLLM variants with ViT3D and \modelname\ as video encoder (the variants that pretrained on 8M video data and 50M image data),  and fine-tune the linear connector using the instruction tuning dataset from LlaVA-video~\cite{zhang2024video}. We train both VideoLLMs for 3 epochs with a batch size of 32 and evaluate the models on various QA benchmarks, including VideoMME~\cite{fu2024video}, NextQA~\cite{xiao2021next}, ActivityNetQA~\cite{yu2019activitynet}, MLVU~\cite{zhou2024mlvu}, TempCompass~\cite{liu2024tempcompass}, and MovieChat~\cite{song2024moviechat}. We follow the setup and evaluation metrics in the LMMs-Eval repository~\cite{lmms_eval2024} for all benchmarks. 
The results are shown in Tab.~\ref{tab:videollm}. \textbf{\modelname-LLM outperforms ViT3D-LLM across all QA benchmarks}, possibly because the modern VideoQA tasks often require reasoning about fine-grained objects and events in video, in which our trajectory-based tokenization helps.
\textbf{\modelname-LLM trains 4.2x faster and has 18x fewer inference FLOPs in average than ViT3D-LLM} because of the significantly fewer vision tokens proposed by our video encoder. 
Additionally, we specifically test how the trained VideoLLM performs in long videos by varying its input number of frames in MovieChat long video benchmark. We find that our model scales significantly better than ViT3D in both inference FLOPs and accuracy (Fig.~\ref{fig:vlm_scaling}), demonstrating its strong potential in building long-term video models.

\subsection{Ablation Study}
\label{sec:ablation}

We conduct extensive ablations on the impact of segmentation and tracking, trajectory generation pipeline, and the trajectory encoder architecture. Due to computational constraints, all models in this study are trained on a randomly sampled 880k subset of Panda-70M, and we use ViT-Base as the transformer backbone.

\subsubsection{Impact of Segmentation and Tracking}
When generating trajectories, our tokenization paradigm relies on segmentation and tracking models, both of which introduce unique inductive biases in generated tokens: segmentation reduces spatial redundancy and adds semantic information, while tracking reduces temporal redundancy. To understand their contributions, we analyze the individual impact of segmentation and tracking on model performance: the segmentation-only baseline segments each frame individually without considering temporal redundancy, while the tracking-only baseline performs tracking on 16x16 patches without segmentation. We use the patch tracking model in~\cite{jabri2020space}. Both baselines use the same tokenizer architecture as \modelname. As shown in Tab.~\ref{tab:inductive_bias}, Both baselines underperform \modelname. The segmentation-only model improves ViT3D on retrieval tasks but underperforms in both classification tasks, perhaps because spatially semantic tokens help the model better capture objects but fail to recognize actions due to the lack of temporal connections. Additionally, we find that patch-based tracking is not robust, as tracking in patches is less semantically meaningful than tracking object segments. This results in the patch-tracking model underperforming \modelname.

\subsubsection{Ablation of Trajectory Generation Pipeline}
We ablate different design choices of the trajectory generation pipeline and show performance in  Tab.~\ref{tab:ablations}.

\noindent\textbf{Segmentation model.}  
The model that uses segments from DirectSAM consistently outperforms the one with segments from SAM, likely due to DirectSAM's ability to produce more detailed segmentation masks.

\noindent\textbf{Trajectory resolution.}  
Both DirectSAM and SAM2 are originally trained at a resolution of 1024. While using their original resolution provides more precise boundary in the generated masks, we observe minimal impact on downstream performance. Given the 4$\times$ reduction in compute, we adopt 512$\times$512 resolution as our default setting.

\noindent\textbf{Frame partitioning strategy.}   
We find that using frame stride only underperforms in retrieval-related tasks, possibly because some objects in the mid-video fail to be detected at their first appearance. Using key frame only degrades performance in the SSV2 classification task. We suspect that this is because object changes rarely occur in SSV2, leading to very few detected key frames. As a result, each token is overloaded to represent lengthy trajectories.

\subsubsection{Ablation of Trajectory Encoder Architecture}
The ablations for different components of trajectory encoder architecture are shown in Tab.~\ref{tab:ablations}.

\noindent\textbf{Feature extractor.}  
Extracting features from ResNet-18 significantly outperform hierarchical transformer such as Hiera-small, perhaps because convolutional filters provide better spatial features, especially for small modules. 

\noindent\textbf{Necessity of hierarchical features.}  
Incorporating hierarchical features improves performance, likely because they help compensate for information loss during mask pooling operations when extracting region-specific features.

\noindent\textbf{Necessity of the position branch.}  
Although the appearance token may already encode some positional information from ResNet features, we find that an explicit position branch remains essential for strong performance.

\noindent\textbf{Feature aggregation across frames.}  
For aggregating per-frame features, we find that using perceiver resampler significantly outperforms simple mean-pooling, highlighting the importance of modeling temporal variation.

\noindent\textbf{Number of query tokens per trajectory.}
Increasing the number of query tokens per trajectory provides only marginal performance improvements. This suggests that trajectory information is already heavily compressed before reaching the perceiver resampler. Addressing this potential bottleneck is left to future work.

\begin{table}[t]
\centering
\resizebox{.8\linewidth}{!}{
  {\def\arraystretch{1.1}
\begin{tabular}{lccc}
  \toprule
   \textbf{Variation} & \textbf{SSV2} & \textbf{K400} & \textbf{Retrieval}  \\
  \midrule
  \textcolor{gray}{\textit{vit3d}} & \textcolor{gray}{\textit{17.2}} & \textcolor{gray}{\textit{21.8}} & \textcolor{gray}{\textit{12.0}} \\
Segmentation-only & 16.4 & 19.9 & 17.0 \\
Tracking-only & 19.3  & 18.1 & 15.4 \\
\modelname & \textbf{22.7} & \textbf{25.9} & \textbf{17.8}\\
  \bottomrule
\end{tabular}}}
\vspace{-1mm}
\caption{{\bf Individual impact of segmentation and tracking}. We can see that removing segmentation (tracking-only) or tracking (segmentation-only) significantly hurts model performance.}
\label{tab:inductive_bias}
\end{table}

\begin{table}[t]
\centering
\resizebox{.85\linewidth}{!}{
  {\def\arraystretch{1.1}
\begin{tabular}{c|ccccc}
  \toprule
   & \textbf{Objectives} & \textbf{Variation} & \textbf{SSV2} & \textbf{K400} & \textbf{Retrieval}  \\
  \midrule
  & \multicolumn{2}{c}{\baseline{\modelname}} & \baseline{22.7} & \baseline{25.9} & \baseline{17.8}\\
  \cmidrule(lr){1-6}
  \multirow{4}{*}{\rotatebox{90}{\textbf{traj-gen}}}
  & Resolution & 1024$\to$512 & 22.5 \textcolor{gray}{{\footnotesize ( \(\downarrow\) 0.2)}} & 26.3 \textcolor{gray}{{\footnotesize ( \(\uparrow\) 0.4)}} & 18.0 \textcolor{gray}{{\footnotesize ( \(\uparrow\) 0.2)}} \\
  \cmidrule(lr){2-6}
  & \multirow{1}{*}{Seg. Type} & D-SAM$\to$SAM & 22.1 \textcolor{gray}{{\footnotesize ( \(\downarrow\) 0.6)}} & 23.5 \textcolor{gray}{{\footnotesize ( \(\downarrow\) 2.4)}} & 16.7 \textcolor{gray}{{\footnotesize ( \(\downarrow\) 1.1)}} \\
  \cmidrule(lr){2-6}
  & \multirow{2}{*}{Clip Par.} 
  & Key frame only & 20.6 \textcolor{gray}{{\footnotesize ( \(\downarrow\) 2.1)}} & 25.7 \textcolor{gray}{{\footnotesize ( \(\downarrow\) 0.2)}} & 17.6 \textcolor{gray}{{\footnotesize ( \(\downarrow\) 0.2)}} \\
  &  & Frame cut only & 22.7 \textcolor{gray}{{\footnotesize (0)}} & 25.3 \textcolor{gray}{{\footnotesize ( \(\downarrow\) 0.6)}} & 17.3 \textcolor{gray}{{\footnotesize ( \(\downarrow\) 0.5)}} \\
  \midrule
  \multirow{6}{*}{\rotatebox{90}{\textbf{traj-encoder}}} 
  & \multirow{1}{*}{Hiera. Feat.} & Yes$\to$No & 22.4 \textcolor{gray}{{\footnotesize ( \(\downarrow\) 0.3)}} & 25.1 \textcolor{gray}{{\footnotesize ( \(\downarrow\) 0.8)}} & 17.5 \textcolor{gray}{{\footnotesize ( \(\downarrow\) 0.3)}} \\
  \cmidrule(lr){2-6}
  & \multirow{1}{*}{Pos. Branch} & Yes$\to$No & 20.6 \textcolor{gray}{{\footnotesize ( \(\downarrow\) 2.1)}} & 25.3 \textcolor{gray}{{\footnotesize ( \(\downarrow\) 0.6)}} & 17.4 \textcolor{gray}{{\footnotesize ( \(\downarrow\) 0.4)}} \\
  \cmidrule(lr){2-6}
  & \multirow{1}{*}{ConvNet} & ResNet$\to$Hiera & 18.3 \textcolor{gray}{{\footnotesize ( \(\downarrow\) 4.4)}} & 21 \textcolor{gray}{{\footnotesize ( \(\downarrow\) 4.9)}} & 16.0 \textcolor{gray}{{\footnotesize ( \(\downarrow\) 1.8)}} \\
  \cmidrule(lr){2-6}
  & \multirow{1}{*}{Feature Aggr.} & Percei.$\to$Mean & 19.2 \textcolor{gray}{{\footnotesize ( \(\downarrow\) 3.5)}} & 23.4 \textcolor{gray}{{\footnotesize ( \(\downarrow\) 2.5)}} & 16.5 \textcolor{gray}{{\footnotesize ( \(\downarrow\) 1.3)}} \\
  \cmidrule(lr){2-6}
  & \multirow{2}{*}{Query Token} & 1$\to$2 & 22.7 \textcolor{gray}{{\footnotesize (0)}} & 26.0 \textcolor{gray}{{\footnotesize ( \(\uparrow\) 0.1)}} & 17.8 \textcolor{gray}{{\footnotesize (0)}} \\
  &  & 1$\to$4 & 22.8 \textcolor{gray}{{\footnotesize ( \(\uparrow\) 0.1)}} & 26.2 \textcolor{gray}{{\footnotesize ( \(\uparrow\) 0.3)}} & 17.9 \textcolor{gray}{{\footnotesize ( \(\uparrow\) 0.1)}} \\
  \bottomrule
\end{tabular}}}
\vspace{-1mm}
\caption{\textbf{Ablation study of both trajectory generation pipeline and trajectory encoder architecture.}}
\label{tab:ablations}
\vspace{-3mm}
\end{table}

\section{Conclusion}

\modelname tokenizes videos using panoptic sub-object trajectories and combines it with a video encoder. \modelname outperforms standard video tokenization in both accuracy and efficiency, while other token-efficient video encoders improve efficiency at the expense of accuracy.

\noindent\textbf{Acknowledgement.} This project was funded by Toyota Motor Inc.

\clearpage
\setcounter{page}{1}
\maketitlesupplementary
\section{More Implementation Details}

We provide the complete training details in Table~\ref{tab:pre-training hyperparams}. 
We optimize the models using AdamW optimizer~\cite{loshchilov2017decoupled} with a learning rate of $10^{-4}$, a weight decay of $10^{-2}$, and mixed precision training. We adopt a cosine annealing learning rate schedule. The contrastive view (batch size) for video training is set to 256, and all models are trained for 30 epochs. We train all models with 32 NVIDIA H100 GPUs. 
For data augmentation, we apply a combination of random ColorJitter, Grayscale, Gaussian blur, horizontal flip, and resized cropping during training. At testing, we use only a simple resizing operation to ensure consistency. 

\noindent\textbf{Hyperparameters for downstream MAP probing.} For two downstream evaluations that require MAP probing,  We use AdamW
optimizer with weight decay 0.5, and set learning rate to be
0.0001. We also layer-normalize the video features before providing them to the classifier. We use a batch size of 128, and we train the classifier for 12 epochs.
\begin{table}[h]
    \centering
    \centering
     \begin{tabular}{l|c}
        \toprule Hyperparameter & Value \\
        \midrule
        Trasformer size & vit-large \\
        Resolution      & 224       \\
        Frame sampling  & uniform 16 frames \\
        Optimizer       & AdamW \\
        Base LR         & $1e^{-4}$ \\
        Weight decay    & 0.02   \\
        Optimizer momentum  & $\beta_1=0.9, \beta_2=0.999$ \\
        Batch size      & video-256, image-4096 \\
        Training epochs & 30 \\
        LR schedule     & cosine decay \\
        Warm up epochs  & 1 \\
        Warm up schedule & linear warm-up \\
        Random crop scale & (0.2, 1.0) \\
        Random crop ratio & (3/4, 4/3) \\
        Horizontal flip probability & 0.5 \\
        Color jitter probability  & 0.8 \\
        Gaussian blur probability & 0.5 \\
        Grayscale probability & 0.2 \\
        \bottomrule 
    \end{tabular}
    \caption{hyperparameters used for pre-training.}
    \label{tab:pre-training hyperparams}
    \end{table}

\begin{table*}[htbp]
    \centering
    \resizebox{0.98\linewidth}{!}{
        \begin{tabular}{c|c|c|c}
         \textbf{Module} & \textbf{Detail} & \textbf{Output Shape} & \textbf{Parameter Size} \\
        \midrule
       Per-frame Feature Extractor & 
        sum
        $\left\{\begin{array}{c}
        \text{ResNet18 stage 1 + linear (64$\to$64) + resize (56$\times$56)} \\ 
        \text{ResNet18 stage 2 + linear (128$\to$64) + resize (28$\to$56)} \\ 
        \text{ResNet18 stage 3 + linear (256$\to$64) + resize (14$\to$56)} \\ 
        \text{ResNet18 stage 4 + linear (512$\to$64) + resize (7$\to$56)}
        \end{array}\right\}$ 
        & $T\times56\times56\times64$ & 11.6M \\
        \midrule
        Mask Pooling & Mask pooling per trajectory (total $N$ trajectories) & $N\times T\times64$ & 0 \\
        \midrule
    Sinusoidal Encoder & bounding box coordinate (4) $\to$ high-dimensional embeddings (64) & $N\times T\times64$ & 0 \\
        
        \midrule
        Perceiver Resampler  & 
        Multi-head cross attention 
        $\left\{\begin{array}{c}
        \text{Query: 1;  Layers: 1} \\ 
        \text{Heads: 8; Dim: 64} \\ 
        \end{array}\right\}$ $\times$ 2
        &  $N\times T\times64$ & 8.4M \\
        \midrule
        MLP  & Linear (64$\to$1024) $\times$ 2 & $N\times1024$ & 0.13M\\
        \midrule
        \textcolor{gray}{\textit{Main Transformer}} & \textcolor{gray}{\textit{Transformer module of ViT-Large}} & \textcolor{gray}{\textit{$N\times1024$}} & \textcolor{gray}{\textit{304M}} \\
        \bottomrule
        \end{tabular}
    }
    \caption{Detailed architecture of our model.}
    \label{tab:arc}
\end{table*}

\begin{figure*}[htbp]
    \centering
    \includegraphics[width=0.9\linewidth]{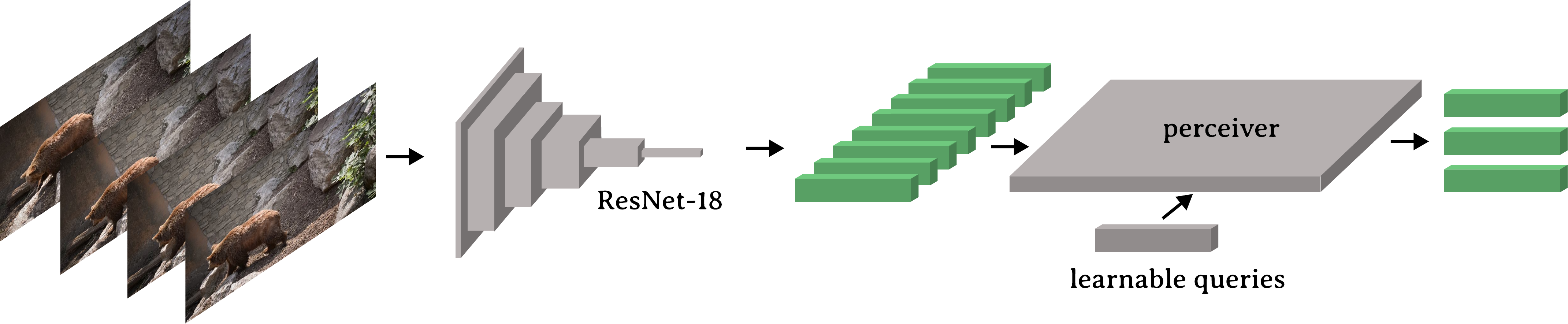}
    \caption{Architecture for TokenMerge baseline.}
    \label{fig:baseline_arc}
\end{figure*}

\begin{table}[h]
\centering
\setlength{\tabcolsep}{6pt}
\begin{tabular}{lccc} 
    \toprule
    \textbf{Model} & \textbf{K400} & \textbf{SSV2} & \textbf{UFC-101} \\
    \midrule
    ViT3D & \underline{42.0} & \textbf{12.3} & \underline{40.4} \\
    TokenLearner & 40.9 & 11.0 & 37.8 \\
    ViViT & 39.9 & 11.5 & 34.8 \\
    \blname & 38.4 & 10.3 & 35.4 \\
    RLT & 41.0 & 10.3 & 33.7 \\
    ToMe & 38.2 & 9.9 & 37.1 \\ 
    \modelname (ours) & \textbf{42.4} & \underline{11.8} & \textbf{42.1} \\
    \bottomrule
\end{tabular}
\vspace{-2mm}
\caption{\textbf{Zero-shot action classification performance.} We report top-5 accuracy.}
\label{tab:zero_shot_classification}
\end{table}

\begin{table*}[th]
\centering
\setlength{\tabcolsep}{8pt}
\begin{tabular}{llcccccccc}
\toprule
\textbf{Training Data} & \textbf{Model} &
\multicolumn{2}{c}{\textbf{ActivityNet}} &
\multicolumn{2}{c}{\textbf{VATEX}} &
\multicolumn{2}{c}{\textbf{MSR-VTT}} &
\multicolumn{2}{c}{\textbf{Charades}} \\
\cmidrule(lr){3-4}
\cmidrule(lr){5-6}
\cmidrule(lr){7-8}
\cmidrule(lr){9-10}
 & & \small txt2vid & \small vid2txt & \small txt2vid & \small vid2txt & \small txt2vid & \small vid2txt & \small txt2vid & \small vid2txt \\
\midrule
\multirow{2}{*}{panda-2m}
 & ViT3D 
   & 26.33 & 27.33 
   & 24.74 & 44.80 
   & 23.75 & 48.30 
   &  7.11 &  7.29 \\
 & \modelname (ours)  
   & 31.97 & 31.97 
   & 28.94 & 51.40 
   & 26.94 & 50.60 
   & 10.14 & 10.47 \\
\midrule
\multirow{2}{*}{panda-4m}
 & ViT3D 
   & 35.34 & 34.54
   & 35.00 & 59.11
   & 30.90 & 56.61
   & 12.61 & 12.61 \\
 & \modelname (ours)  
   & 38.62 & 38.41
   & 36.19 & 61.02
   & 31.71 & 60.52
   & 14.81 & 14.81 \\
\midrule
\multirow{2}{*}{panda-8m}
 & ViT3D 
   & 38.82 & 37.46
   & 40.59 & 64.46
   & 34.71 & 60.83
   & 17.45 & 16.00 \\
 & \modelname (ours)  
   & 42.35 & 41.92
   & 41.35 & 65.35
   & 35.22 & 62.73
   & 19.41 & 18.36 \\
\bottomrule
\end{tabular}
\vspace{-2mm}
\caption{\textbf{Full retrieval performance for pretraining video data scaling experiment.} We report results on four commonly used video retrieval datasets for both text-to-video (\texttt{txt2vid}) and video-to-text (\texttt{vid2txt}).}
\label{tab:data_scaling}
\end{table*}

\begin{table*}[th]
\centering
\setlength{\tabcolsep}{5pt}
\begin{tabular}{llcccccccc}
\toprule
\textbf{Training Data} & \textbf{Model} &
\multicolumn{2}{c}{\textbf{ActivityNet}} &
\multicolumn{2}{c}{\textbf{VATEX}} &
\multicolumn{2}{c}{\textbf{MSR-VTT}} &
\multicolumn{2}{c}{\textbf{Charades}} \\
\cmidrule(lr){3-4} \cmidrule(lr){5-6} \cmidrule(lr){7-8} \cmidrule(lr){9-10}
 & & \small txt2vid & \small vid2txt & \small txt2vid & \small vid2txt & \small txt2vid & \small vid2txt & \small txt2vid & \small vid2txt \\
\midrule
\multirow{2}{*}{panda8m} 
 & ViT3D & 37.82 & 34.54 & 39.59 & 59.11 & 33.71 & 56.51 & 15.35 & 12.61 \\
 & \modelname (ours)  & 41.35 & 38.41 & 40.35 & 61.02 & 34.22 & 61.00 & 18.41 & 14.81 \\
\midrule
\multirow{2}{*}{panda8m + datacomp50m}
 & ViT3D & 43.62 & 44.65 & 47.16 & 70.70 & 41.16 & 68.74 & 21.25 & 20.50 \\
 & \modelname (ours)  & 53.57 & 53.36 & 50.38 & 75.10 & 47.38 & 79.96 & 24.80 & 22.01 \\
\bottomrule
\end{tabular}
\vspace{-2mm}
\caption{\textbf{Full retrieval performance for incorporating image data experiment.} We report R@5 scores on four commonly used video retrieval datasets. \texttt{txt2vid} is text-to-video retrieval and \texttt{vid2txt} is video-to-text retrieval.}
\label{tab:video_image}
\end{table*}

\begin{table}[h]
\centering
\setlength{\tabcolsep}{2pt}
\begin{tabular}{lccc} 
    \toprule
    \textbf{Model} & \multicolumn{1}{c}{\textbf{ImageNet}} & \multicolumn{2}{c}{\textbf{COCO}} \\
    \cmidrule(lr){2-2} \cmidrule(lr){3-4}
    metrics & Top-5 Acc & img2txt R@5 & txt2img R@5 \\
    \midrule
    ViT & 77.7 & 73.6 & 58.3 \\
    \modelname (ours) & 74.9 & 71.1 & 55.5 \\
    \bottomrule
\end{tabular}
\vspace{-2mm}
\caption{\textbf{Performance for image-only experiments.} We report top-5 accuracy for ImageNet classification and Recall@5 for COCO retrieval (image-to-text \& text-to-image).}
\label{tab:image_only}
\end{table}

\section{More Architecture Details}

To complement the main paper, we provide additional details on our model architecture and TokenMerge baseline's architecture.

\noindent\textbf{Trajectory Encoder.}  
We provide the complete architectural details of our trajectory tokenizer in table~\ref{tab:arc}. As shown,  the parameter size of our tokenizer is an order of magnitude smaller compared with main transformer.

\noindent\textbf{TokenMerge Baseline.}  
Although the size of our trajectory encoder is very small (20M) compared with the transformer encoder (304M), to ensure that our improvements do not simply come from adding parameters, we train a model that uses exactly the same modules as \modelname but uses a learnable token merging mechanism that does not incorporate trajectory priors.
The architecture of the TokenMerge baseline is illustrated in Figure~\ref{fig:baseline_arc}. We design it such that the only difference from our trajectory tokenizer is whether it incorporates trajectory priors when compressing tokens. All other architectural modules remain identical to ensure a controlled comparison. The output token number is set to be 1024 to match the average FLOPs of our model at training set (including trajectory generation). 

\section{Key Frame Detection Algorithm}
We illustrate the details of our key frame detection algorithm, which ensembles three sub-detectors to ensure robust scene boundary identification. A frame is classified as a key frame if it is proposed by at least two out of the three detectors. All detectors are implemented using the Content-Aware Detector from the PySceneDetect package.

\noindent\textbf{HSV Colorspace Detector.}  
This detector operates in the HSV color space. Each frame is converted from RGB to HSV, and the average difference across all channels is computed frame by frame. A scene change is triggered if the difference between adjacent frames exceeds threshold 27.

\noindent\textbf{Luminance Histogram Detector.}  
Each frame is converted from its original color space to YCbCr, and the Y channel (luminance) is extracted. The normalized histogram of the Y channel in the current frame is then compared to that of the previous frame using the correlation method (cv2.HISTCM\_CORREL). A scene change is detected if the histogram correlation between consecutive frames falls below a set threshold 0.15.

\noindent\textbf{RGB Detector.}  
This detector computes an intensity value for each frame by averaging the R, G, and B values across all pixels, yielding a single floating-point number. A scene cut is triggered if the intensity difference between consecutive frames exceeds threshold 12.

\section{Detailed setup in AVAv2 Spatial Temporal Detection task}
We follow the setup in~\cite{tong2022videomae} to evaluate our model on the AVAv2 spatial-temporal action detection task. In this task, given an object's bounding box in a specific video frame, the model must predict the action associated with that object at that time instant. This requires extracting video features corresponding to the region of interest (ROI) and applying a probing head to classify the action based on the localized features. We use the same attentive probing head across all models, but the ROI pooling strategy differs depending on each model’s tokenization mechanism, which we illustrate below:

\noindent\textbf{ViT3D.}  
ViT3D produces a spatial-temporal feature map, allowing us to use ROIAlign to extract features corresponding to the given bounding box. This setup is the same as~\cite{tong2022videomae}

\noindent\textbf{Our Model.}  
Since each output token in our model corresponds to an object trajectory, we leverage its segmentation mask at each timestep to determine its presence within the bounding box. We gather all tokens whose trajectories have at least 80\% of their segmentation mask area inside the bounding box at the annotated frame.

\noindent\textbf{ViViT.}  
ViViT is a two-stage model, where the first stage outputs spatial features, and the second stage extracts temporal features. We handle this by pooling its spatial features using ROIAlign and selecting the corresponding temporal feature based on the annotated timestep. The final feature is obtained by concatenating the pooled spatial and temporal representations.

\noindent\textbf{TokenLearner.}  
We use the TokenFuser module proposed in its original paper to reproject pruned tokens back to their original spatial locations. This allows us to perform feature pooling in the same manner as ViT3D.

\noindent\textbf{RLT \& ToMe.}  
Both RLT and ToMe dynamically merge space-time patch tokens that are identified as redundant. We reassemble the feature map by duplicating merged features back to their corresponding redundant patches, then ROI pool the reconstructed feature map in the same way as ViT3D.

\noindent\textbf{TokenMerge.}  
Since TokenMerge learns to merge tokens in a fully data-driven manner, it does not retain explicit spatial correspondences to the original input grid. As a result, we are unable to pool features corresponding to the region of interest.

\section{Full tables for scaling performance experiments}
We provide the complete table for the scaling up experiments, which we only show the plots of average trend in the main table. Table~\ref{tab:data_scaling} presents the performance variations of the model with the change of the scale of the training data. Table~\ref{tab:video_image} presents the model's performance with images adding to training data.

\section{Zero-shot action classification}
We report zero-shot action classification performance for all models here as a complement for attentive probing action classification that shown in the main paper. We note that for video model, attentive probing that only involves vision encoder is a more accurate measure for action classification task,  because the text template for action is hard to construct and likely to be out-of-distribution for text that model saw during training (e.g. put something on something).  Nevertheless, as shown in table~\ref{tab:zero_shot_classification}, our model still has competitive performance under zero-shot setting, outperforming most of baseline models.

\section{Visualizations of generated trajectories}
\begin{figure*}[htbp]
    \centering
    \includegraphics[width=0.9\linewidth]{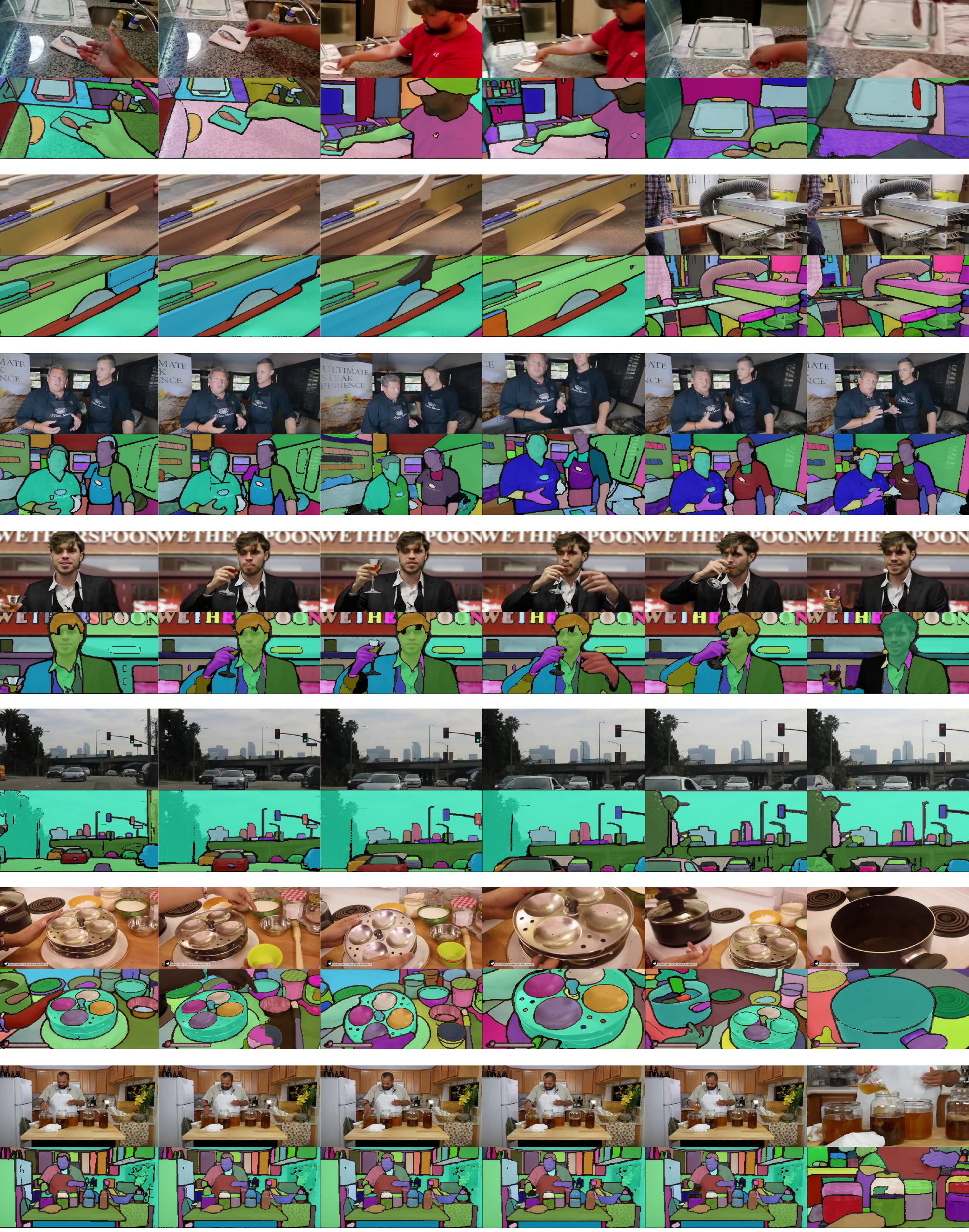}
    \caption{\textbf{Visualizations of our generated trajectories (part 1).}}
    \label{fig:vis1}
\end{figure*}

\begin{figure*}[htbp]
    \centering
    \includegraphics[width=0.9\linewidth]{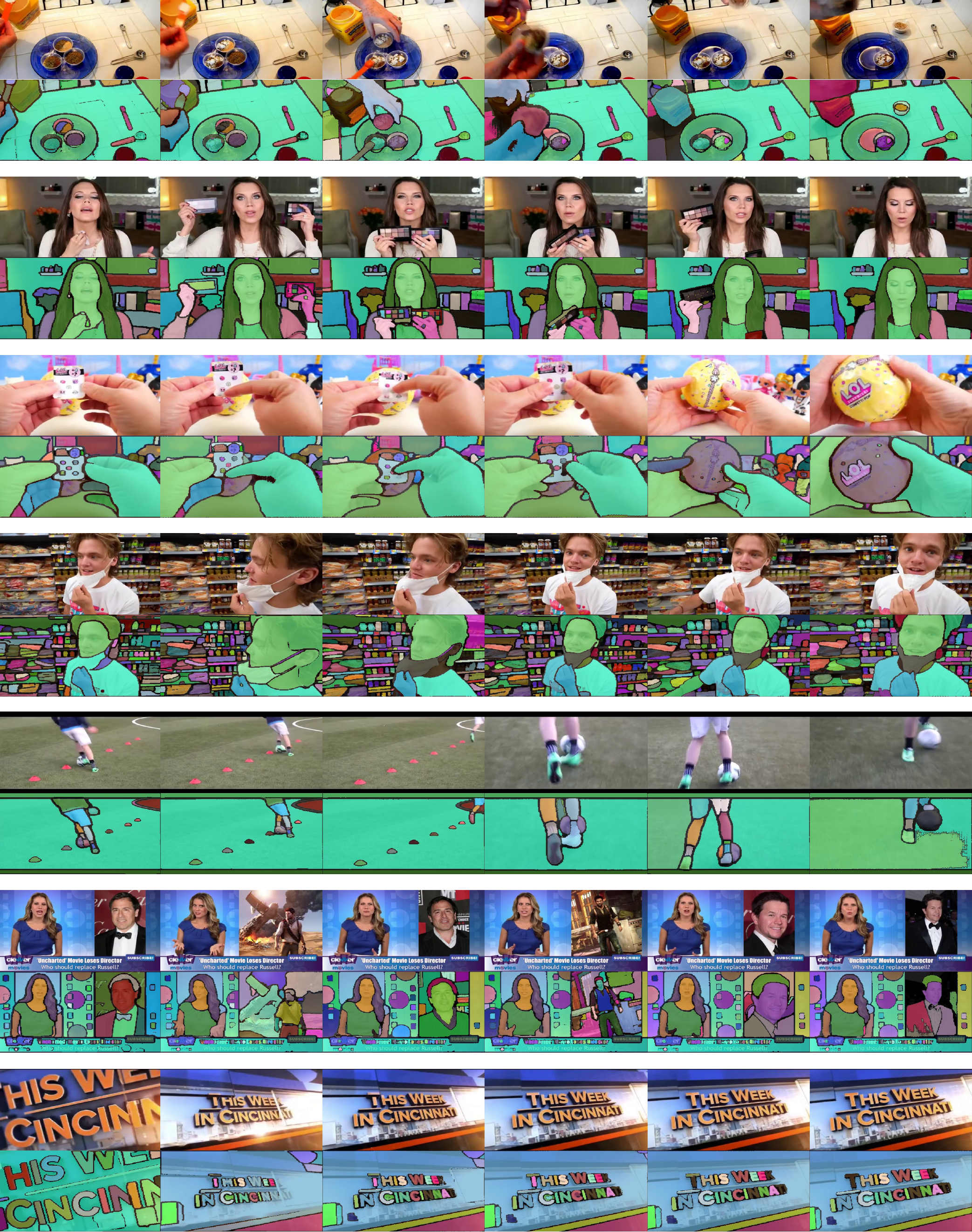}
    \caption{\textbf{Visualizations of our generated trajectories (part 2).}}
    \label{fig:vis2}
\end{figure*}

We show examples of generated trajectories in our training set at figure~\ref{fig:vis1} and figure~\ref{fig:vis2}. Our pipeline allows us to generate high-quality panoptic trajectory with high efficiency. The generated segments are in detailed subobject level, allowing us to reason fine-grained interaction. The tracking is also very robust attributing to the powerful SAM2 model.
We do observe occasional matching failure for the same objects between sub-clips (like frame 3$\to$ 4 in example 3), causing the same object being split into multiple trajectories.

\section{Image only experiments}
Since we mention our model can naturally be adapted to image data, it will be interesting to see its performance in image-only training as well. We therefore train our model at datacomp50M image-captioning dataset, and compare it to regular ViT model that trains in the same dataset (table~\ref{tab:image_only}. We found our model underperforms ViT in downstream evaluation of ImageNet zero-shot classification and COCO zero-shot image-text retrieval. This means our bigger gain in incorperating image data experiment for our model is primarily because our model can train at image and video  together, avoiding image-then-video pipeline and the information loss when transferring 2D model's weight to 3D model. How to improve our model design to let it become also competitive in image-only domain is left to future work.


{
    \small
    \bibliographystyle{ieeenat_fullname}
    \bibliography{main}
}
\end{document}